\documentclass[runningheads,a4paper]{llncs}

\usepackage[margin=1.23in]{geometry}
\usepackage{graphicx}
\usepackage[sectionbib]{natbib}
\usepackage{multirow}
\usepackage{amsmath}         
\usepackage{amssymb}  
\usepackage{url}
\usepackage{multirow}
\usepackage{setspace}

\newcommand{\keywords}[1]{\par\addvspace\baselineskip
\noindent\keywordname\enspace\ignorespaces#1}

\long\def\symbolfootnote[#1]#2{
\begingroup
	\def\thefootnote{\fnsymbol{footnote}}\footnote[#1]{#2}
\endgroup}

\begin{document}

\mainmatter  

\title{Introducing Quantum-Like Influence Diagrams for Violations of the Sure Thing Principle}
\titlerunning{Lecture Notes in Computer Science}

\author{Catarina Moreira$^1$ \and Andreas Wichert$^2$ }

\institute{$^1$School of Business, University of Leicester\\ University Road, LE1 7RH, Leicester, United Kingdom\\
\email{cam74@le.ac.uk}\\
~\\
$^2$Instituto Superior T\'{e}cnico, INESC-ID\\Av. Professor Cavaco Silva, 2744-016 Porto Salvo, Portugal\\
 \email{andreas.wichert@tecnico.ulisboa.pt  }
}

\maketitle

\begin{abstract}

It is the focus of this work to extend and study the previously proposed quantum-like Bayesian networks~\citep{Moreira14, Moreira16} to deal with decision-making scenarios by incorporating the notion of maximum expected utility in influence diagrams. The general idea is to take advantage of the quantum interference terms produced in the quantum-like Bayesian Network to influence the probabilities used to compute the expected utility of some action. This way, we are not proposing a new type of expected utility hypothesis. On the contrary, we are keeping it under its classical definition. We are only incorporating it as an extension of a probabilistic graphical model in a compact graphical representation called an influence diagram in which the utility function depends on the probabilistic influences of the quantum-like Bayesian network.

Our findings suggest that the proposed quantum-like influence digram can indeed take advantage of the quantum interference effects of quantum-like Bayesian Networks to maximise the utility of a cooperative behaviour in detriment of a fully rational defect behaviour under the prisoner's dilemma game.

\end{abstract}

\section{Introduction}

In this work, we extend the Quantum-Like Bayesian Network previously proposed by~\citet{Moreira14,Moreira16} by incorporating the framework of expected utility. This extension is motivated by the fact that quantum-like models tend to explain the probability distributions in several decision scenarios where the agent (or the decision-maker) tends to act irrationally~\citep{Busemeyer12book,Busemeyer15trends}. By irrational, we mean that an individual chooses strategies that do not maximise or violate the axioms of expected utility. It is not enough to know these probability distributions. On the contrary, it would be desirable to use this probabilistic information to help us act upon a real world decision scenario. For instance, if a patient has cancer, it is not enough for a doctor to know the probability distribution of success of different treatments. The doctor needs act and choose a treatment based on specific information about the patient and how this treatment will affect him/her. Probabilistic models are used in tasks that reason under uncertainty, in other words, they are models that reach a conclusion based on partial evidence. Decision-making models such as the expected utility hypothesis, are used to decide how to act in the world. The main problem with such decision-making models is that it is very challenging to determine the right action in a decision task where the outcomes of the actions are not fully determined~\citep{koller09prob}. For this reason, we suggest to extend the previously proposed Quantum-Like Bayesian Network to a Quantum-Like Influence diagram where we take into account both the quantum-like probabilities (incorporating quantum interference effects) of the various outcomes and the preferences of an individual between these outcomes.

Generally speaking, an Influence diagram is a compact directed acyclical graphical representation of a decision scenario originally proposed by~\citet{Howard84} which consists in three types of nodes: random variables (nodes) of a Bayesian Network, action nodes representing a decision that we need to make, and an utility function. The goal is to make a decision, which maximises the expected utility function by taking into account probabilistic inferences performed on the Bayesian Network. However, since influence diagrams are based on classical Bayesian Networks, then they cannot cope with the paradoxical findings reported over the literature. 

It is the focus of this work to study the implications of incorporating Quantum-Like Bayesian Networks in the context of influence graphs. By doing so, we are introducing quantum interference effects that can disturb the final probability outcomes of a set of actions and affect the final expected utility. We will study how one can use influence diagrams to explain the paradoxical findings of the prisoner’s dilemma game based on expected utilities.

\section{Revisiting the Prisoner's Dilemma and the Expected Utility Hypothesis}

\begin{figure}[h!]
\centering
\includegraphics[scale=0.5]{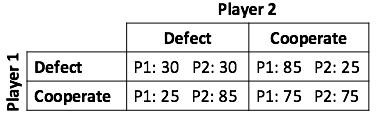}
\caption{Example of a payoff matrix used in the \citet{Shafir92} Prisoner's Dilemma experiment}
\label{fig:payoff}
\end{figure}

The Prisoner's Dilemma game consists in two players who are in two separate confinements with no means of communicating with each other. They were offered a deal: if one defects against the other, he is set free while the other gets a heavy charge. If they both defect, they get both a big charge and if they both cooperate by remaining silent, they get a small charge. Figure~\ref{fig:payoff} shows an example of a payoff matrix for the Prisoner's Dilemma used in the experiments of~\citet{Shafir92} where the goal is to score the maximum number of points.

Looking at the payoff matrix, one can see that the best action for $both$ players is to $cooperate$, however experimental findings show that the majority of the players choose to $defect$ even when it is known that the other player chose to $cooperate$. The Prisoner's Dilemma is a clear example of how two perfectly rational individuals choose to defect (they prefer an individual reward), rather than choosing the option which is best for both (to cooperate). The expected utility hypothesis is a framework that enables us to explain why this happens.

The expected utility hypothesis corresponds to a function designed to take into account decisions under risk. It consists of a choice of a possible set of actions represented by a probability distribution over a set of possible payoffs \citep{Neumann53}. It is given by Equation~\ref{eq:EU},

\begin{equation}
EU = \sum_i Pr( x_i ) \cdot U(x_i),
\label{eq:EU}
\end{equation}
where $U( x_i )$ is an utility function associated to event $x_i$.

In the experiment of~\citet{Shafir92}, the participant needed to choose between de actions $defect$ or $cooperate$. We will address to this participant as player 2, $P2$ and his opponent, to player 1, $P1$. According to the expected utility hypothesis, $P2$ would have to choose the action that would grant him the highest expected utility. Assuming that we do not know what $P1$ chose (so we model this with a neutral prior of 0.5), we can compute the expected utility of Player 2 as
\[ EU[ Defect ] = 0.5 \times U( P1 = D, P2 = D ) + 0.5 \times U( P1 = C, P2 = D ) =  57.5,\]
\[ EU[ Cooperate] = 0.5 \times U( P1 = D, P2 = C ) + 0.5 \times U( P1 = C, P2 = C ) = 50. \]
Note that $U( P1 = x, P2 = y )$ corresponds to the utility of player 1 choosing action $x$ and player 2 choosing action $y$. The calculations show that the action that maximises the player's expected utility is  $Defect$. This is what it is known as the {\it Maximum Expected Utility hypothesis} (MEU).

In the end of the 70's, Daniel Kahneman and Amos Tversky showed in a set of experiments that in many real life situations, the predictions of the expected utility were completely inaccurate \citep{Tversky74,Kahneman82book,Kahneman79}. This means that a decision theory should be predictive in the sense that it should say what people actually {\it do choose}, instead of what they {\it must choose}. The Prisoner's Dilemma game is one of the experiments that show the inaccuracy of the expected utility hypothesis by showing violations to the laws of classical probability. Table~\ref{tab:pd} summarises the results of several words of the literature reporting violations to the total law of classical probability. All of these works tested three conditions in the Prisoners Dilemma Game: (1) the player knows the other defected ( {\it Known to Defect}), (2) the player knows the other cooperated ({\it Known to Collaborate}), (3) the player does not know the other player's action ({\it Unknown}). This last condition shows a deviation from the classical probability theory, suggesting that there is a significant percentage of players who are not acting according to the maximum expected utility hypothesis. 

\begin{table}[h!]
\resizebox{\columnwidth}{!} {
\begin{tabular}{l|c|c|c|c}
{\bf Literature	}			& {\bf Known to Defect} 	& {\bf Known to Collaborate} 	& {\bf Unknown}	& {\bf Classical Probability} \\
\hline
\citet{Shafir92}					& 	0.9700			&	0.8400				& 0.6300			& 0.9050 \\
\citet{Li02} (Average)				&	0.8200			&	0.7700				& 0.7200			& 0.7950\\
\citet{Li02} {\bf Game 1}			& 0.7333 			& 0.6670 					& 0.6000			& 0.7000						\\
\citet{Li02} {\bf Game 2}			& 0.8000			& 0.7667					& 0.6300			& 0.7833						\\
\citet{Li02} {\bf Game 3}			& {\bf 0.9000}		& {\bf 0.8667} 				& {\bf 0.8667}		& {\bf 0.8834}					\\
\citet{Li02} {\bf Game 4}			& 0.8333			& 0.8000					& 0.7000			& 0.8167						\\
\citet{Li02} {\bf Game 5}			& 0.8333			& 0.7333					& 0.7000			& 0.7833						\\
\citet{Li02} {\bf Game 6}			& {\bf 0.7667}		& {\bf 0.8333}				& {\bf 0.8000}		& {\bf 0.8000}					\\
\citet{Li02} {\bf Game 7}			& {\bf 0.8667}		& {\bf 0.7333}				& {\bf 0.7667}		& {\bf 0.8000}					\\
\hline

\end{tabular}
}
\caption{Works of the literature reporting the probability of a player choosing to defect under several conditions. The entries of the table that are highlighted correspond to experiments where the violations of the sure thing principle were not found.}
\label{tab:pd}
\end{table}

Table~\ref{tab:pd} presents several examples where the  principle of maximum expected utility is not, in general, an adequate descriptive model of human behaviour. In fact, people are often irrational, in the sense that their choices do not satisfy the principe of maximum expected utility relative to any utility function~\citep{koller09prob}.

Previous works in the literature have proposed quantum-like probabilistic models that try to accommodate these paradoxical scenarios and violations to the Sure Thing Principle~\citep{busemeyer06,Busemeyer09markov,Moreira17lestah,Busemeyer09,Busemeyer12book}. There is also a vast amount of work in trying to extend the expected utility hypothesis to a quantum-like versions~\cite{Mura09,Yukalov15}. However, the expected utility framework alone poses some difficulties, since it is very challenging the task of decision-making in situations where the outcomes of an action are not fully determined~\citep{koller09prob}. 

In this paper, we try to fill this gap by taking into account the quantum-like probability inferences produced by a quantum-like Bayesian network to various outcomes and extend these probabilities to influence the preferences of an individual between these outcomes. Note that the probabilistic inferences produced by the quantum-like Bayesian network will suffer quantum interference effects in decision scenarios under uncertainty. The general idea is to use these quantum interference effects to influence the expected utility framework in order to favour other actions than what would be predicted from the classical theory alone. We will combine this structure in a directed and acyclic compact probabilistic graphical model for decision-making, which we will define as the quantum-like influence diagram.

\section{A Quantum-Like Influence Diagram for Decision-Making}

A Quantum-Like Influence Diagram is a compact directed acyclical graphical representation of a decision scenario, which was originally proposed by~\citet{Howard84}. It consists on a set of random variables $X_1, \dots, X_N$ belonging to a quantum-like Bayesian network. Each random variable $X_i$ is associated with a conditional probability distribution (CPD) table, which describes the distribution of quantum probability amplitudes of the random variable $X_i$ with respect to its parent nodes, $\psi( X_i | Pa_{X_i})$. Note that the difference between a quantum-like Bayesian network and a classical network is simply the usage of complex numbers instead of classical real numbers. The usage of complex numbers will enable the emergence of quantum interference effects. The influence diagram also consists in an utility node defined variable $U$, which is associated with a deterministic function $U( Pa_U)$. The goal is to make a decision, which maximises the expected utility function by taking into account probabilistic inferences performed on the quantum-like Bayesian network.

\begin{figure}[h!]
\centering
\includegraphics[scale=0.25]{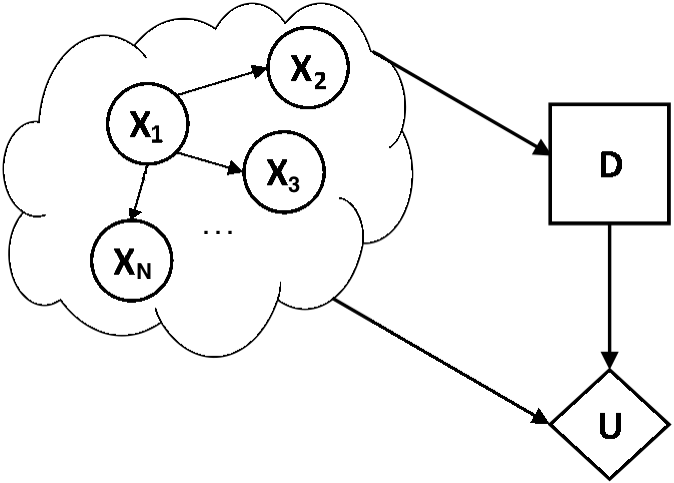}
\caption{General example of a Quantum-Like Influence Diagram comprised of a Quantum-Like Bayesian Network, $X_1, ..., X_N$, a Decision Node, $D$, and an Utility node with no children, $U$.}
\label{fig:influence}
\end{figure}

An example of a quantum-like influence diagram is presented in Figure~\ref{fig:influence}. In the Figure, one can notice the three different types of nodes: (1) random variable nodes (circle-shape), denoted by $X_1, \cdots, X_N$, of some Quantum-Like Bayesian Network, (2) a decision node (rectangle-shape), denoted by $D$, which corresponds to the decision that we want to make, and (3) an Utility node (diamond-shape), denoted by $U$, which in the scope of this paper, will represent the payoffs in the Prisoner's Dilemma Game. 

The goal is to maximise the expected utility by taking into consideration the probabilistic inferences of the quantum-like bayesian network, which makes use of the quantum interference effects to accommodate and predict violations to the Sure Thing Principle.

In the next sections, we will address each of these three components separately. 

\section{Quantum-Like Bayesian Networks}

Quantum-like Bayesian Network have been initially proposed by~\citet{Moreira14, Moreira16, Moreira18} and they can be defined by a directed acyclic graph structure in which each node represents a different quantum random variable and each edge represents a direct influence from the source node to the target node. The graph can represent independence relationships between variables, and each node is associated with a conditional probability table that specifies a distribution of quantum complex probability amplitudes over the values of a node given each possible joint assignment of values of its parents. In other words, a quantum-like Bayesian Network is defined in the same way as classical network with the difference that real probability values are replaced by complex probability amplitudes.

In order to perform exact inferences in a quantum-like Bayesian network, one needs to compute the:

\begin{itemize}

\item {\bf Quantum-Like full join probability distribution.} The quantum-like full joint complex probability amplitude distribution over a set of $N$ random variables $\psi(X_1, X_2, ..., X_N)$ corresponds to the probability distribution assigned to all of these random variables occurring together in a Hilbert space. Then, the full joint complex probability amplitude distribution of a quantum-like Bayesian Network is given by: 

\begin{equation}
\psi( X_1, \dots, X_N ) =  \prod_{j = 1}^N \psi ( X_j | Parents( X_j ) ) 
\label{eq:joint_q}
\end{equation}
Note that, in Equation~\ref{eq:joint_q}, $ X_i $ is the list of random variables (or nodes of the network), $Parents(X_i)$ corresponds to all parent nodes of $X_i$ and $\psi\left( X_i \right)$ is the complex probability amplitude associated with the random variable $X_i$. The probability value is extract by applying Born's rule, that is, by making the squared magnitude of the joint probability amplitude, $\psi \left( X_1, \dots, X_N \right)$:
\begin{equation}
Pr( X_1, \dots, X_N ) = \left| \psi( X_1, \dots, X_N ) \right|^2
\end{equation}

\item {\bf Quantum-Like Marginalization.} Given a query random variable $X$ and let $Y$ be the unobserved variables in the network, the marginal distribution of $X$ is simply the amplitude probability distribution of $X$ averaging over the information about $Y$. The quantum-like marginal probability for discrete random variables, can be defined by Equation~\ref{eq:inference_q}. The summation is over all possible $y$, i.e., all possible combinations of values of the unobserved values $y$ of variable $Y$. The term $\gamma$ corresponds to a normalisation factor. Since the conditional probability tables used in Bayesian Networks are not unitary operators with the constraint of double stochasticity (like it is required in other works of the literature~\citep{busemeyer06,Busemeyer09}), we need to normalise the final scores. This normalisation is consistent with the notion of normalisation of wave functions used in Feynman's Path Diagrams.

In classical Bayesian inference, on the other hand, normalisation is performed due to the independence assumptions made in Bayes rule. 

\begin{equation}
\begin{split}
Pr(  X | e ) = \gamma \left|~ \sum_y \prod_{k=1}^N \psi( X_k | Parents(X_k), e, y ) ~\right| ^2
\end{split}
\label{eq:inference_q}
\end{equation}
Expanding Equation~\ref{eq:inference_q}, it will lead to the quantum marginalisation formula~\citep{Moreira14}, which is composed by two parts: one representing the classical probability and the other representing the quantum interference term (which corresponds to the emergence of destructive / constructive interference effects):
\begin{equation}
Pr(X | e) = \gamma \sum_{i = 1}^{|Y|}  \left| \prod_k^N  \psi( X_k | Parents(X_k), e, y = i )  \right| ^2 + 2 \cdot {\mathit Interference}
\label{eq:bn_inference_q}
\end{equation}
\begin{multline*}
{\mathit Interference} =\\  \sum_{i=1}^{|Y|-1} \sum_{j=i+1}^{|Y|}  \left| \prod_k^N \psi( X_k | Parents(X_k), e, y=i ) \right| \cdot \left| \prod_k^N  \psi( X_k | Parents(X_k), e, y= j ) \right| \cdot \cos( \theta_i - \theta_j )
\end{multline*}

\end{itemize}

Note that, in Equation~\ref{eq:bn_inference_q}, if one sets $(\theta_i - \theta_j)$ to $\pi/2$, then $\cos( \theta_i - \theta_j) = 0$. This means that the quantum interference term is canceled and the quantum-like Bayesian Network collapses to its classical counterpart.

Formal methods to assign values to quantum interference terms are still an open research question, however some work has already been done towards that direction~\citep{Yukalov11,Moreira16,Moreira17faces}. 

\section{Maximum Expected Utility in Classical Influence Diagrams}


Given a set of possible decision rules, $\delta_A$, the goal of Influence Diagrams is to compute the decision rule that leads to the Maximum Expected Utility. 

\begin{equation}
EU\left[ \mathcal{D}\left[\delta_A \right] \right] = \sum_x Pr_{\delta_A}\left( x | a \right) U \left( x , a \right)
\label{eq:EU_init}
\end{equation}

The goal is to choose some action $a$ that maximises the expected utility:

\[ a^* = argmax_{\delta_A} EU\left[ \mathcal{D} \left[ \delta_A\right] \right] \]

One can map the expected utility formalism to the scope of Bayesian networks in the following way. In the expected utility formula, $Pr_{\delta_a}( x | a )$ corresponds to a full joint probability distribution of all possible outcomes, $x$, given different actions $a$. This means that we can decompose the full joint probability distribution to the chain rule of probability theory as the product of each node with its parent nodes.
\begin{equation}
EU\left[ \mathcal{D}\left[\delta_A \right] \right] = \sum_x Pr_{\delta_A}\left( x | a \right) U \left( x , a \right)~~~~~~~~~~~~~~~~~~~~~~~~~~~~~~~~~~~~~~~
\end{equation}

\begin{equation}
EU\left[ \mathcal{D}\left[\delta_A \right] \right] = \sum_{X_1, \dots, X_n, A}  \left(  \left( \prod_{i} Pr\left( X_i | Pa_{X_i} \right)  \right) U\left( Pa_U\right) \delta_A \left( A | Z \right)  \right)
\label{eq:fact1}
\end{equation}

In Equation~\ref{eq:fact1}, $Z = Pa_A$ represents the parent nodes of action $A$. We can factorise Equation~\ref{eq:fact1} in terms of the decision rule, $\delta_A$, obtaining
\begin{equation}
 EU\left[ \mathcal{D}\left[\delta_A \right] \right] = = \sum_{Z,A} \delta_A \left( A | Z \right) \left( \sum_W  \left(  \prod_i Pr\left( X_i | Pa_{X_i} \right) \right) U\left( Pa_U \right) \right),
\end{equation}
where $W = \{ X_1, \dots, X_N \} - Z$ corresponds to all nodes of the Bayesian Network that are not contained in the set of nodes in $Z$.

By marginalising the summation over $W$, we obtain an expected utility formula that is written only in terms of the factor $\mu (A,Z)$. Note that this factor corresponds to a conditional distribution table of random variable $Z$ (the outcomes of some action $a$) and action $a$.

\begin{equation}
EU\left[ \mathcal{D}\left[\delta_A \right] \right] =  \sum_{Z, A}  \delta_A\left( A | Z \right) \mu \left( A, Z \right)
\end{equation}

The Maximum Expected Utility for a classical Influence Diagrams is given by~\citep{koller09prob}:

\begin{equation}
\delta^*_A \left( a, Z \right) =  \alpha(x)=\left\{
                \begin{array}{c c}
                  1	& ~~~~~~~~~~a = argmax\left( A, Z \right)\\
                  0 	& otherwise\\
                \end{array}
              \right.
\end{equation}

\section{Maximum Expected Utility in Quantum-Like Influence Diagrams}

The proposed quantum-like influence diagram is built upon the formalisms of quantum-like Bayesian networks. This means that real classical probabilities need to be replaced by complex quantum amplitudes. 
 
We start the derivation with the initial notion of expected utility already presented in the previous section.

\begin{equation}
EU\left[ \mathcal{D}\left[\delta_A \right] \right] = \sum_x Pr_{\delta_A}\left( x | a \right) U \left( x , a \right)
\label{eq:EU_init}
\end{equation}

For simplicity, let's consider the decision scenario where we have two binary events $X_1$ and $X_2$. Then, we can decompose the classical expected utility equation as

\begin{equation}
EU\left[ \mathcal{D}\left[\delta_A \right] \right] = \sum_{X_1, X_2, A}  \delta_A (A | X_1 )Pr \left( X_1 \right) Pr\left( X_2 | X_1 \right) U \left( X_1 , A \right)
\end{equation}

Like before, we can factorise this formula in terms of the decision rule $\delta_A$, obtaining

\begin{equation}
EU\left[ \mathcal{D}\left[\delta_A \right] \right] 	= \sum_{A, X_2}  \delta_A (A | X_2 ) \sum_{X_1} Pr \left( X_1 \right) Pr\left( X_2 | X_1 \right) U \left( X_1 , A \right)
\end{equation}

For binary events, we obtain the marginalisation of $X_1$ over both $X_2$ and $D$

\begin{equation}
	EU\left[ \mathcal{D}\left[\delta_A \right] \right]  = \sum_{A, X_2}  \delta_A (A | X_2 ) \cdot \mu \left( X_2, A \right)
\label{eq:qEU_init}
\end{equation}

where $\mu \left( X_2, A \right)$  is a factor with the utility function expressed in terms of the distribution of $X_1$. More specifically, it is given by
\begin{equation}
\begin{split}
\mu \left( X_2, A \right) = Pr \left( X_1 = t \right) Pr\left( X_2 | X_1 = t \right) U \left( X_1 = t , A \right) +\\
				 	+ Pr \left( X_1 = f \right) Pr\left( X_2 | X_1 = f \right) U \left( X_1 = f , A \right)
\end{split}
\label{eq:qEU_exp}
\end{equation}

Since the proposed quantum-like influence diagram makes use of a quantum-like Bayesian network, this means that we need to convert the classical real probabilities into complex quantum amplitudes. This is performed by applying Born's rule: for some classical probability $A$, the corresponding quantum amplitude is simply its squared magnitude, $Pr(A) = \left| \psi_A \right|^2$~\citep{Deutsch88,Zurek11}. Since in Equation~\ref{eq:qEU_exp} we have a combination of utility functions with probability values, we cannot apply Born's rule directly, since we would not be satisfying its definition. For this reason, we propose to split Equation~\ref{eq:qEU_exp} into a vector representation containing a classical probability and another containing the utility function. This procedure is similar to the one propose in the Quantum Decision Theory model of~\citet{Yukalov15} where the authors separate a prospect into an utility factor (a factor containing the classical utility of a lottery) and an attraction factor (a probabilistic factor that results from the quantum interference effect).

Considering $\pi_a$ the vector representation of a classical probability vector and $u_a$ the classical utility corresponding to the choice of some action $A$, then we obtain
\[ \pi_a =  Pr \left( X_1 = t \right) Pr\left( X_2 | X_1 = t \right) + Pr \left( X_1 = f \right) Pr\left( X_2 | X_1 = f \right)  \]
\[ u_a =  U \left( X_1 = t , A \right) + U \left( X_1 = f , A \right) \] 

We can apply Born's rule by replacing classical real numbers by quantum-like amplitudes and performing their squared magnitude as
\begin{equation}
\begin{split}
\pi_a = \left| q_a \right|^2 \Leftrightarrow  q_a = \left|  \psi(X_1=t) \psi( X_2 | X_1 = t) + \psi(X_1=f) \psi(X_2 | X_1 = f) \right|^2~~~~~~~~~~~~~~~~ \\
q_a =  \left|  \psi(X_1=t) \psi( X_2 | X_1 = t ) \right|^2 + \left| \psi(X_1=f) \psi( X_2 | X_1 = f ) \right|^2 + Interf,
\end{split}
\end{equation}
where the quantum interference term is given by
\begin{equation}
Interf = 2 \left|  \psi(X_1=t) \psi( X_2 | X_1 = t ) \right| \left| \psi(X_1=f) \psi( X_2 | X_1 = f ) \right| Cos\left( \theta_1 - \theta_2 \right).
\end{equation}

The utility factor $u_a$ needs to be updated in order to become a factor of the quantum interference term. 
\[  u_a =  U \left( X_1 = t , A \right) + U \left( X_1 = f , A \right) +  U \left( X_1 = t , A \right) \cdot U \left( X_1 = f , A \right)\] 

The result of this marginalisation, $\mu \left( X_2, A \right) $, will be given by the product of the vector representation of these two terms:
\[ \mu \left( X_2, A \right) =  \langle q_a | u_a \rangle, \]
where the vector representation corresponds to
\[ 
| q_a \rangle = \left[ \begin{array}{c}
				 \left|  \psi(X_1=t) \psi( X_2 | X_1 = t ) \right|^2 \\
				 \left|  \psi(X_1=f) \psi( X_2 | X_1 = f ) \right|^2\\ 
				Interf 
				\end{array}\right]~~~~~~~~~~
| u_a \rangle = \left[ \begin{array}{c}
				 U \left( X_1 = t , A \right) \\
				 U \left( X_1 = f , A \right) \\ 
				U \left( X_1 = t , A \right) U \left( X_1 = f , A \right)
				\end{array}\right].
\]
This way, the final marginalisation for the quantum-like influence diagram is
\begin{equation}
\begin{split}
 \mu \left( X_2, A \right) = \langle q_a | u_a \rangle = ~~~~~~~~~~~~~~~~~~~~~~~~~~~~~~~~~~~~~~~~~~~~~~~~~~~~~~~~~~~~~~~~~~~~~ \\ 
	\left|  \psi(X_1=t) \psi( X_2 | X_1 = t ) \right|^2 \cdot U \left( X_1 = t , A \right) + \left|  \psi(X_1=f) \psi( X_2 | X_1 = f ) \right|^2 \cdot U \left( X_1 = f , A \right) + \dots \\
	+ Interf \cdot U \left( X_1 = t , A \right) U \left( X_1 = f , A \right)
\end{split}
\label{eq:final}
\end{equation}

Note that, in Equation~\ref{eq:final}, if one sets the interference term $(\theta_i - \theta_j)$ to $\pi/2$, then $\cos( \theta_i - \theta_j) = 0$. This means that the quantum interference term is canceled and the quantum-like influence diagram collapses to its classical counterpart. In other words, one can see the quantum-like influence diagram as a more general and abstract model of the classical diagram, since it represents both classical and quantum behaviour.

Finally, the Maximum Expected Utility for Quantum-Like Influence Diagrams is given by:
\begin{equation}
\delta^*_A \left( a, Z \right) =  \alpha(x)=\left\{
                \begin{array}{c c}
                  1	& ~~~~~~~~~~a = argmax~\mu \left(  X_2, A  \right)\\
                  0 	& otherwise~~~\\
                \end{array}
              \right.
\end{equation}

\section{A Quantum-Like Influence Diagram for the Prisoner's Dilemma Game}

Several paradoxical findings have been reported over the literature showing that individuals do not act rationally in decision scenarios under uncertaint~\citep{Kuhberger01,Tversky92,Lambdin07,Hristova08,Busemeyer06proceed}.. 

The quantum-like influence diagram can help to accommodate and explain the several paradoxical decisions by manipulating the quantum interference effects that emerge from the inferences in the quantum-like Bayesian network. These inferences can then be used to reestimate the expected utility of an agent.

\begin{figure}[h!]
\centering
\includegraphics[scale=0.45]{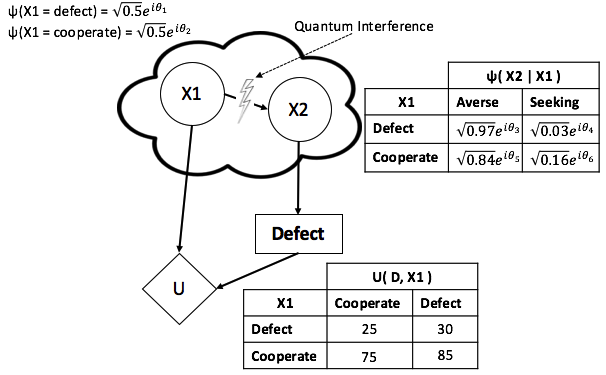}
\caption{Quantum-Like Infleunce Diagram representing the Prisoner's DIlemma Experiment from~\citep{Shafir92}. }
\label{fig:pd}
\end{figure}

We will model the works previously reported in Table~\ref{tab:pd} under the proposed quantum-like influence diagram. Figure~\ref{fig:pd} corresponds to the representation of the work of~\citet{Shafir92}. The three types of nodes in the represented quantum-like influence diagram are the following:

\begin{itemize}
\item {\bf Random Variables:} the circle-shaped nodes are the random variables belonging to the quantum-like bayesian network representing the player that needs to make a decision in the Prisoner's Dilemma, without being aware of the decision of his opponent. We modelled this network with two binary random variables, $X_1$ and $X_2$. $X_1$ corresponds to the player's own strategy (either to defect or to cooperate) and $X_2$ the player's personal risk preferences, i.e. either he is risk averse (and therefore would find it safe to engage in a $defect$ strategy) or he is risk seeking (and would prefer to engage in a $cooperate$ strategy). The tables next to each random variable are conditional probability tables and they show the probability distribution of the variable towards its parent nodes. These conditional probability tables match the probability distributions reported in Table~\ref{tab:pd}. In the specific case of Figure~\ref{fig:pd}, this table is filled with the values of the probability amplitudes identified in the work of~\citet{Shafir92}. For the general case, we will assume that the player has no initial strategy and we will assume neutral priors for the variable $X_1$ (like it was assumed in previous works of the literature, see~\cite{Moreira16}). \\*

\item {\bf Action Node:} the rectangle shaped node is the action that we want to make a decision. In the context of the prisoner's dilemma we are interested to compute the maximum expected utility of defecting or not defecting (i.e. cooperating).\\*

\item {\bf Utility Node:} the diamond shaped node corresponds to the payoffs that the player will have for taking (or not) the action $defect$, given his own personal preferences towards risk. The values in this node will be populated with the different payoffs used across the different experiments of the prisoner's dilemma reported over the literature.
\end{itemize}

In the conditions where the player $knows$ the strategy of his opponent, the quantum-like influence diagram collapses to its classical counterpart, since there is no uncertainty. This was already noticed in the previous works of~\citet{Moreira14,Moreira16,Moreira17faces}. However, when the player is not informed about his opponent's decision, then the quantum-like Bayesian network will produce interference effects (Equation~\ref{eq:bn_inference_q}). When computing the maximum expected utility, we will marginalise out $X_1$ like it was shown in Equation~\ref{eq:qEU_exp}. This will result in a factor showing the distribution of the player's personal preferences towards risk (either risk averse or risk seeking) towards his actions (either to defect or cooperate). The quantum interference term will play an important role to determine which quantum parameters can influence the player's decision to switch from a classical (and rational) defect action towards the paradoxical decision found in the works the literature, i.e. to cooperate.

\begin{figure}[h!]
\resizebox{\columnwidth}{!} {
\includegraphics{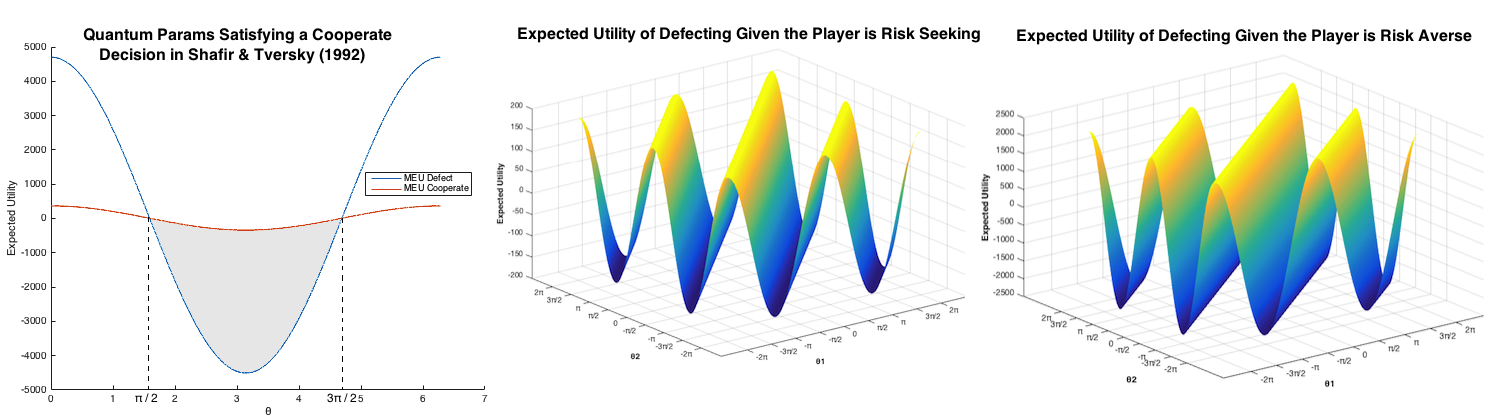}
}
\caption{Impact of quantum interference terms in the overall expected utility: (left) quantum parameters that maximize a cooperate decision, (center) variation of the expected utility when the player is risk averse and (right) variation of the expected utility when the player is risk seeking.}
\label{fig:prelim_res}
\end{figure}

Figure~\ref{fig:prelim_res} demonstrates the impact of the quantum interference effects in the player's decision. The graphs in the centre and in the right of Figure~\ref{fig:prelim_res} represent all possible maximum expected utilities that the player can achieve by varying the quantum interference term $\theta$ in Equation~\ref{eq:final} for a personal preference of being risk averse or risk seeking, respectively. On the left of Figure~\ref{fig:prelim_res}, it is represented all the values of $\theta$ that satisfy the condition that $EU[Cooperate] > EU[Defect]$, i.e., all the values of the quantum interference parameter $\theta$ that will maximise the utility of cooperation rather than defect. One can note that, for experiment of~\citet{Shafir92} (as well as in the remaining works of the literature analysed in this work), one can  maximise the expected utility of Cooperation when the utilities are negative. This is in accordance with the previous study of~\cite{Moreira16} in which the authors found that violations to the Sure Thing Principle imply destructive (or negative) quantum interference effects. As we will see in the next section, the quantum parameters found that are used to maximise the expected utility of a cooperate action lead to destructive quantum interferences and can exactly explain the probability distributions observed in the experiments.

\subsection{Results and Discussion}

\begin{figure}[h!]
\parbox{.4\linewidth}{
	\centering
	\includegraphics[scale=0.3]{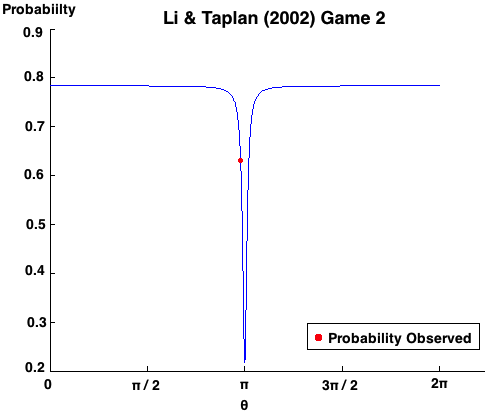}
	\caption{Probabilities that can be obtained in Game 2 of~\citet{Li02}.}
	\label{fig:chaos_1}	
	}
	\hfill
	\parbox{.4\linewidth}{
	\centering
	\includegraphics[scale=0.3]{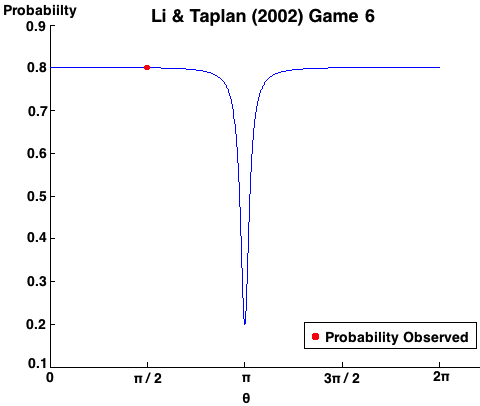}
	\caption{Probabilities that can be obtained in Game 2 of~\citet{Li02}.}
	\label{fig:chaos_2}	
	}
\end{figure}

Although there are several quantum parameters that satisfy the relationship that shows that participants can maximise the utility of a cooperate action, only a few parameters are able to accommodate both the paradoxical probability distributions reported in the several works in the literature and to maximise the expected utility of cooperating. For instance, Figures~\ref{fig:chaos_1} and~\ref{fig:chaos_2} show how the quantum parameters are sensitive to accommodate the violations of the Sure Thing Principle in terms of the probability distributions. The slight variation of the quantum parameter $\theta$ in the quantum-like Bayesian network can lead to completely different probability distributions which differ from the ones observed in the difference experimental scenarios reported in the literature. These probability distributions will influence the utilities computed by the expected utility framework.

In Table~\ref{tab:li_theta}, it is presented the quantum parameters that lead to the quantum interference term that is necessary to fully explain and accommodate the violations to the sure thing principle reported over several works of the literature.

\begin{table} [h!]
\resizebox{\columnwidth}{!} {
\begin{tabular}{l | c | c | c | c | c | c | c | c   }
				& \citet{Shafir92}	& \citet{Li02} G1	& \citet{Li02} G2 & \citet{Li02} G3	& \citet{Li02} G4	& \citet{Li02} G5 & \citet{Li02} G6 & \citet{Li02} G7  \\
\hline				
{\bf Prob of Defect}		& 			& 			& 			& 			& 			& 			& 			& 			\\
(Known to Defect) 		& 0.9700		& 0.7333		& 0.8000		& {\bf 0.9000}	& 0.8333		& 0.8333		& {\bf 0.7667}	& {\bf 0.8667}		\\
\hline
{\bf Prob of Cooperate}	& 			& 			& 			& 			& 			& 			& 			& 			\\		
(Known to Cooperate) 	& 0.8400		& 0.6670		& 0.7667		& {\bf 0.8667}	& 0.8000		& 0.7333		& {\bf 0.8333}	& {\bf 0.7333}		\\
\hline
{\bf Classical Prob}		& 			& 			& 			& 			& 			& 			& 			& 			\\		
(Unknown condition)	& 0.9050		&  0.7000		& 0.7833		& {\bf 0.8834}	& 0.8167		& 0.7833		& {\bf 0.8000}	& {\bf 0.8000}		\\
\hline
{\bf Experim Prob}		& 			& 			& 			& 			& 			& 			& 			& 			\\			
(Unknown condition)	& 0.6300		& 0.6000		& 0.6300		& {\bf0.8667}	& 0.7000		& 0.7000		& {\bf 0.8000}	& {\bf 0.7667}		\\
\hline
{\bf Quantum Interference}	&		&			&			&			&			&			&			&			\\			
$\theta$ param			& 2.8151 	& 3.0170 		& 3.0758		& {\bf 2.8052}	& 3.2313		& 2.8519		& {\bf 1.5708}	& {\bf 3.7812}	\\

\hline
\end{tabular}
}
\caption{Experimental results reported for the Prisoner's Dilemma game. The entries highlighted correspond to games that are not violating the Sure Thing Principle.}
\label{tab:li_theta}
\end{table}

For this reason, we decided to test if the quantum-like parameters used to accommodate the violations to the Sure Thing Principle were sufficient and if they could also lead to a maximisation of expected utility of cooperation. We performed simulations of the different works in the literature and we concluded that the quantum interference effects that can accommodate violations to the violations of the Sure Thing Principle in the quantum-like Bayesian network alone, also explain a higher preference of the cooperative action over defect. Table~\ref{tab:final_res} presents the results.

\begin{table}[h!]
\resizebox{\columnwidth}{!} {
\begin{tabular}{ l | cc | cc | cc | cc | cc | cc | cc | cc |}
\cline{2-17}
	&\multicolumn{2}{|c|}{~}	&\multicolumn{14}{|c|}{\citet{Li02} }\\
	& \multicolumn{2}{|c| }{\citet{Shafir92}} 	&\multicolumn{2}{|c|}{Game 1}	&\multicolumn{2}{|c|}{Game 2}	&\multicolumn{2}{|c|}{Game 3}	&\multicolumn{2}{|c|}{Game 4}		&\multicolumn{2}{|c|}{Game 5}	&\multicolumn{2}{|c|}{Game 6}	&\multicolumn{2}{|c|}{Game 7}	\\
\cline{2-17}
	& {\bf MEU}  	& {\bf MEU} 	& {\bf MEU } 	& {\bf MEU}	& {\bf MEU}  	& {\bf MEU}	& {\bf MEU}  	& {\bf MEU} 	& {\bf MEU}  	& {\bf MEU}	& {\bf MEU } 	& {\bf MEU}	& {\bf MEU}  	& {\bf MEU}	& {\bf MEU}  	& {\bf MEU}\\
	& {\bf (coop)}	& {\bf (def)}	& {\bf (coop)}	& {\bf (def)}	& {\bf (coop)}	& {\bf (def)}	& (coop)	& (def)	& (coop)	& (def)	& (coop)	& (def)	 & MEU  	& MEU	& (coop)	& (def)\\
\cline{1-17}
	CL Rk Av   		& 43.63		& {\bf 50.25}	& 34.19	& {\bf 39.35}	& 38.75	& {\bf 61.78} 	& 26.85	& {\bf 50.33}	& 65.70	& {\bf 67.33}	& 16.27	& {\bf 34.50}	& 17.58	& {\bf 36.50}	& 16.43	& {\bf 35.00}	\\
CL Rk Sk   		& ~6.38		& {\bf ~7.25}	& 15.82	& {\bf 18.15}	& 11.25	& {\bf 17.22} 	& 3.65	& {\bf 26.85}	& 14.80	& {\bf 15.17}	& 5.23	& {\bf 10.5}	& 3.92	& {\bf 8.50}	& 5.07	& {\bf 10.00}	\\
\hline
QL Rk Av   		& {\bf -1559.46}	& -2129.94	& -1263.63 	& -1730.21 	& -1422.69 	& -4787.28 	& -702.24 	& -2075.58 	& -5198.14 	& -5462.41	& -221.05		& -1313.94	& 28.83	& {\bf 36.49}	& -184.75		& -1116.33\\
QL Rk Sk   		& {\bf ~116.66}		& -160.08		& -538.62		& -735.89		& -392.89		& -1320.22 	& -94.44		& -270.75 	& -1162.55	& -1221.47	&-61.44		&  -353.22 	& 3.91	& {\bf 8.50}	& -44.86		& -262.30\\
\hline
QL Interf			& ~	   & ~		& ~	& ~		& ~	& ~		& ~	& ~		& ~	& ~	& ~	& ~	& ~	& ~	& ~	& ~	\\		
$\theta_1 - \theta_2$	& 2.815 & 2.815		& 3.017	& 3.017		& 3.0758	& 3.0758		& 2.805	& 2.805		& 3.23	& 3.23	& 2.8519	& 2.8519	 & 1.5708	& 1.5708	& 3.78	& 3.78	\\
\hline
Payoff				& ~	& ~		& ~	& ~		& ~	& ~		& ~	& ~		& ~	& ~		& ~	& ~		& ~	& ~			& ~	& ~	 \\
dd dc				& 30	& 25	 	& 30	& 25		& 73	& 25		& 30	& 25		& 80	& 78		& 43	& 10		& 30	& 10			& 30	& 10	 \\
cd cc				& 85	& 75		& 85	& 75		& 85	& 75		& 85	& 36		& 85	& 83		& 85	& 46		& 60	& 33	 		& 60	& 33	\\
\hline

\end{tabular}
}
\caption{Results obtained after performing inferences in the quantum-like influence diagram for different works of the literature reporting violations of the Sure Thing Principle in the Prisoner's Dilemma Game. One can see that the Maximum Expected Utility (MEU) was changed to favour a Cooperate strategy using the quantum interference effects of the Quantum-Like Bayesian Network. In the payoffs, $d$ corresponds to $defect$ and $c$ to cooperate. The first payoff corresponds to player 1 and the second to player 2.}
\label{tab:final_res}
\end{table}

In Table~\ref{tab:final_res}, we present the MEU computed for each work in the literature using the classical approach for the different personal preferences of the individual towards risk: either risk seeking ({\it CL Rk Sk}) or risk averse ({\it CL Rk Av}). The classical MEU shows that the optimal strategy is to $defect$ even if the individual has a risk seeking personality (who would be willing to bet on a $cooperate$ action). Of course these results go against the experimental works of the literature which say that a significant percentage of individuals engaged in cooperative strategies.

In opposition, when we use the quantum-like influence diagram, we take advantage of the quantum interference terms that will disturb the probabilistic outcomes of the quantum-like Bayesian networks. Since the utility function depends on the outcomes of the quantum-like Bayesian network, then it is straightforward that quantum interference effects influence indirectly the outcomes of the MEU allowing us to favour a different strategy predicted by the classical MEU.

It is interesting to notice that indeed the parameters used accommodate the violations of the Sure Thing Principle alone in the quantum-like Bayesian Network could also be used to maximise the utility of a Cooperate action. This was verified in all works of the literature analysed except in Game 6 in the work of \citet{Li02}. The reason is that Game 6 is not even reporting a violation to the Sure Thing Principle and could be explained by the classical theory with a minor error percentage. So, if it can be explained under the classical theory, then of course it also tends to favour a $defect$ action over a $cooperate$ one.

\section{Conclusion}

In this work, we proposed an extension of the quantum-like Bayesian Network initially proposed by~\citet{Moreira14, Moreira16} into a quantum-like influence diagram. Influence diagrams are designed for knowledge representation. They are a directed acyclic compact graph structure that represents a full probabilistic description of a decision problem by using probabilistic inferences performed in Bayesian networks~\citep{koller09prob} together with a fully deterministic utility function. Currently, influence diagrams have a vast amount of applications. They can be used to determine the value of imperfect information on both carcinogenic activity and human exposure~\citep{Howard05}, the are used to detect imperfections in manufacturing and they can even be used for team decision analysis~\citep{Detwarasiti05}, valuing real options~\citep{Lander01}, etc.

Although we are aware that more studies need to be conducted in this direction, the preliminary results obtained in this study show that the quantum-like Bayesian network can be extended to deal with decision-making scenarios by incorporating the notion of maximum expected utility in influence diagrams. The general idea is to take advantage of the quantum interference terms produced in the quantum-like Bayesian network to influence the probabilities used to compute the expected utility. This way, we are not proposing a new type of expected utility hypothesis. On the contrary, we are keeping it under its classical definition. We are only incorporating it as an extension of a quantum-like probabilistic graphical model where the utility node depends only on the probabilistic inferences of the quantum-like Bayesian network.

This notion of influence diagrams opens several new research paths. One can incorporate different utility nodes being influenced by different random variables of the quantum-like Bayesian Network. This way one can even explore different interference terms affecting different utility nodes, etc. We plan to carry on with this study and further develop these ideas in future research.

\section{Acknowledgements}

This work was supported by national funds through Funda\c{c}\~{a}o para a Ci\^{e}ncia e a Tecnologia (FCT) with reference UID/CEC/50021/2013. The funders had no role in study design, data collection and analysis, decision to publish, or preparation of the manuscript.

The authors would like to thank Mr. Sebastian Schmidt and Dr. Godfrey Charles-Cadogan for all the discussions about the topics of this work. Their wise comments were truly helpful for the development of the ideas in this manuscript.

\scriptsize
\bibliographystyle{frontiersinSCNS_ENG_HUMS} 

\end{document}